\documentclass[letterpaper, 10 pt, conference]{ieeeconf}  

\IEEEoverridecommandlockouts                              

\title{\LARGE \bf
Boosting Zero-Shot VLN via Abstract Obstacle Map-Based Waypoint Prediction with TopoGraph-and-VisitInfo-Aware Prompting
}

\author{Boqi Li\textsuperscript{*}, Siyuan Li\textsuperscript{*}, Weiyi Wang, Anran Li, Zhong Cao, and Henry X. Liu\textsuperscript{\dag},~\IEEEmembership{Senior Member,~IEEE}
\thanks{This work was supported by the DARPA TIAMAT Challenge. (HR0011-24-9-0429)}
\thanks{Boqi Li, Zhong Cao, Henry X. Liu are with the Department of Civil and Environmental Engineering; Siyuan Li and Weiyi Wang are with the Department of Computer Science and Engineering; Anran Li is with the Department of Robotics, University of Michigan, Ann Arbor, MI, USA.  (e-mails: \tt\small [boqili, syuanlee, wweiyi, anranli, zhcao, henryliu]@umich.edu)}
\thanks{$^{*}$: These authors contributed equally to this work.}
\thanks{\dag: Corresponding author.}
}
\usepackage{textcomp}
\usepackage{url}
\usepackage{verbatim}
\usepackage{svg} 
\usepackage{cite}
\usepackage[hidelinks]{hyperref}
\usepackage{amsmath, amssymb, mathrsfs}
\usepackage[ruled,vlined,linesnumbered]{algorithm2e}
\usepackage{booktabs}
\usepackage{makecell}

\begin{document}

\maketitle
\thispagestyle{empty}
\pagestyle{empty}

\begin{figure*}[t!]
    \centering
    \includegraphics[width=0.95\linewidth]{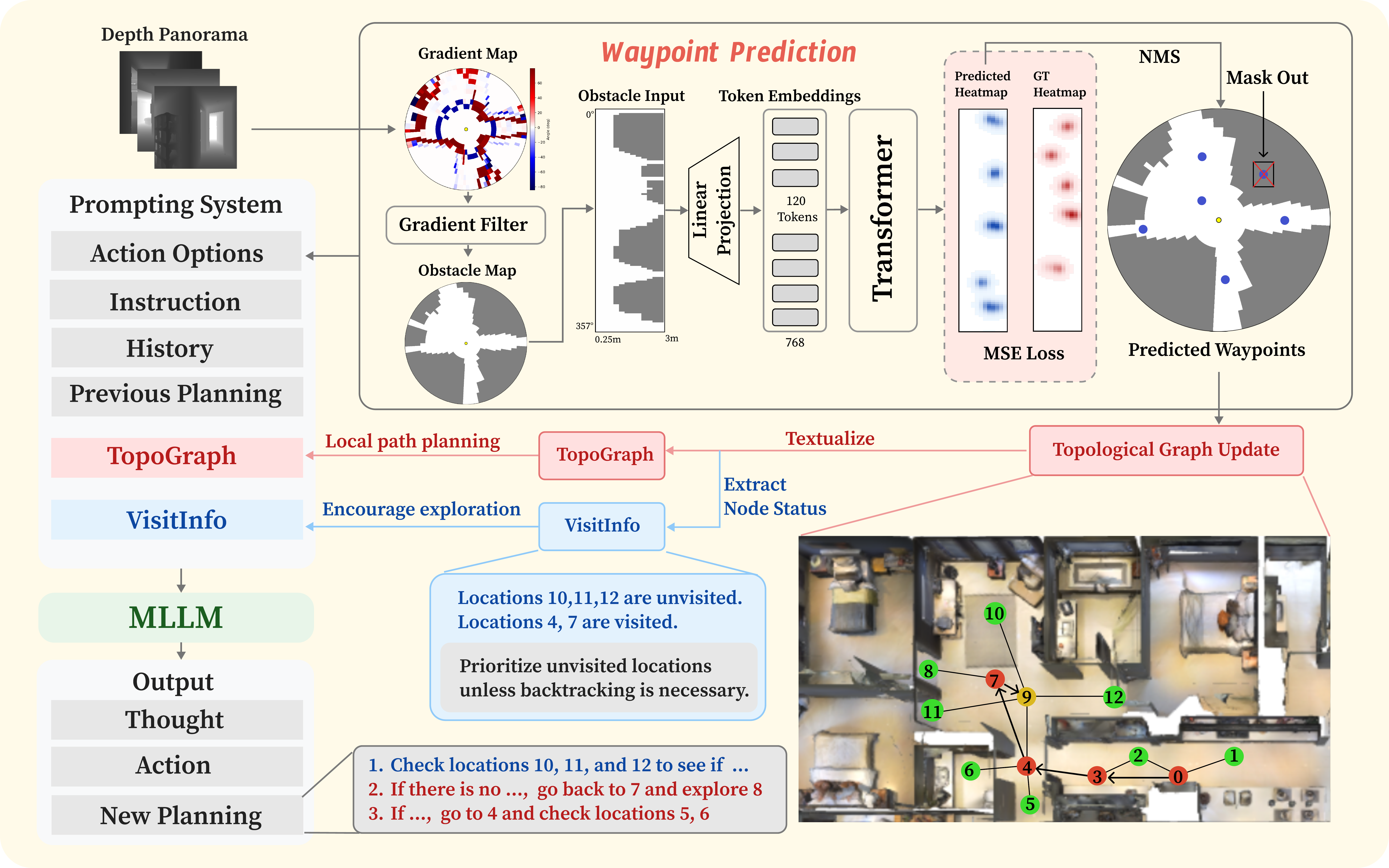}
    \caption{Complete pipeline of our proposed zero-shot VLN with MLLM. Depth images are processed into gradients and filtered to form an obstacle map, which serves as input to the waypoint predictor. Generated waypoints are dynamically added to a topological graph and textualized as TopoGraph and VisitInfo. VisitInfo helps MLLM prioritize exploration and TopoGraph equips MLLM with local path planning (backtracking one/multiple steps and re-exploration) for error correction.}
    \label{pipeline}
\end{figure*}


\begin{abstract}
With the rapid progress of foundation models and robotics, vision-language navigation (VLN) has emerged as a key task for embodied agents with broad practical applications. We address VLN in continuous environments, a particularly challenging setting where an agent must jointly interpret natural language instructions, perceive its surroundings, and plan low-level actions. We propose a zero-shot framework that integrates a simplified yet effective waypoint predictor with a multimodal large language model (MLLM). The predictor operates on an abstract obstacle map, producing linearly reachable waypoints, which are incorporated into a dynamically updated topological graph with explicit visitation records. The graph and visitation information are encoded into the prompt, enabling reasoning over both spatial structure and exploration history to encourage exploration and equip MLLM with local path planning for error correction. Extensive experiments on R2R-CE and RxR-CE show that our method achieves state-of-the-art zero-shot performance, with success rates of 41\% and 36\%, respectively, outperforming prior state-of-the-art methods.

\end{abstract}

\section{INTRODUCTION} 
Vision-language navigation (VLN) is the task in which an embodied agent follows a natural language instruction to navigate through an environment and reach a specified destination. It represents an important capability for embodied AI, with applications ranging from search and rescue to autonomous navigation and daily human–robot interaction. However, VLN remains highly challenging: the agent must not only comprehend natural language instructions but also ground them in its surrounding environment \cite{chang2017matterport3d, r2r, habitat}. Existing approaches \cite{mapgpt, etpnav, constraint} typically rely on RGB-D sensory inputs, which demand both a detailed understanding of complex visual scenes and robust reasoning over the task context.

Early research on VLN has largely focused on discrete environment settings, where a navigation graph is predefined and the agent’s movement is restricted to a finite set of checkpoints \cite{chang2017matterport3d, r2r, rxr}. More recently, it has been extended to continuous environments, where the agent can move freely \cite{habitat,r2rce}. This setting is considerably more challenging and has motivated a line of work that employs waypoint prediction models \cite{Krantz_2021_ICCV}: candidate waypoints are first generated in the continuous space, and a navigation policy then selects one to pursue. This decomposition is intended to ease the learning process by separating spatial grounding from long-horizon reasoning. 

Despite their promise, existing waypoint prediction approaches exhibit drawbacks. First, the choice of input representations remains underexplored. The original waypoint model relied on RGB and depth images; however, subsequent work found that RGB inputs can sometimes degrade performance, while others introduced increasingly complex architectures for RGB-D processing, further increasing model complexity \cite{etpnav, smartway}. Second, predicted waypoints are not guaranteed to be reachable from the agent’s current position and may be obstructed by intervening obstacles. 

Rather than designing waypoint prediction models with increasingly complex sensory inputs and neural architectures, we propose a simplified waypoint prediction model that proves rather effective. Our key insight is that providing raw RGB and depth images introduces large amounts of irrelevant information, which can obscure the underlying spatial relationships and hinder generalization. To address this, we abstract the input representation: depth images are first processed into an obstacle map centered on the agent, which is then used as the sole input to the waypoint predictor. This abstraction encourages the model to focus on the dynamic patterns of obstacle distribution, enabling it to better identify critical regions where waypoints should be generated. Empirically, we find that our approach produces waypoints that are both more feasible and reliable than those generated by prior methods.

We integrate our proposed waypoint model with a zero-shot navigator based on a multimodal large language model (MLLM) \cite{openai2025gpt5}. Zero-shot VLN with large language models (LLMs) has attracted increasing attention in recent years \cite{navgpt,mapgpt,a2nav,opennav,aoplanner}, as purely learning-based navigation models still suffer from data scarcity and struggle to generalize to unseen environments. By leveraging the broad knowledge encoded in LLMs, zero-shot approaches enable agents to better interpret both natural language instructions and perceptual inputs.

Recent works in VLN-CE have explored textual and visual prompt design for LLM-based navigators \cite{a2nav, aoplanner, smartway}, incorporating different forms of contextual information. However, most of these approaches overlook the structure of the topological graph, causing the LLM agent to quickly lose track of which waypoints have already been explored. To address this limitation, we propose a dynamically updated topological graph that explicitly encodes node visitation status. This graph is incorporated into the prompt design, providing the LLM with structured knowledge of explored and unexplored regions, alongside perceptual inputs, thereby enabling more consistent and informed reasoning about the agent’s navigation choices.

We evaluate our proposed waypoint prediction + MLLM navigation framework in the R2R-CE \cite{r2rce} and RxR-CE \cite{rxr} datasets, achieving 41\% and 36\% success rates and outperforming recent zero-shot methods. Our contributions are threefold:
\begin{itemize}
\item Abtract obstacle-map based waypoint prediction: A lightweight predictor using obstacle maps as input, improving waypoint feasibility and reachability.
\item TopoGraph-and-VisitInfo aware prompting: A dynamically updated topological graph with visited-node tracking, incorporated into prompts to help the MLLM reason over spatial structure and exploration history.
\item Integrated zero-shot VLN framework: A unified system combining our waypoint predictor with an MLLM navigator, achieving state-of-the-art results on both R2R-CE and RxR-CE datasets.
\end{itemize}

\section{Related work}
\subsection{Vision-Language Navigation}
Early research in VLN focused on discrete settings. Research on discrete VLN spans imitation and reinforcement learning \cite{Liu_2023_ICCV,wang2024vision,wang2024discovering}, pretraining strategies \cite{DUET,wang2023scaling}, and more recently, zero-shot methods with LLMs \cite{Qiao_2023_ICCV,mapgpt,navgpt,discussnav,zhan2024mc}, collectively laying the foundation for benchmark evaluation and methodological advances in VLN.

Despite progress in discrete VLN, models in this setting remain limited, as agents are restricted to predefined navigation graphs that oversimplify the task. To bridge the gap toward real-world navigation, Savva et al. introduced Habitat \cite{habitat}, a flexible 3D simulator enabling continuous movement and Krantz et al. proposed the more challenging VLN-CE \cite{r2rce}, where agents must also handle low-level control. Numerous approaches have been explored, including waypoint prediction models \cite{dcvln,krantz2020beyond,etpnav}, hierarchical planning with subgoals \cite{irshad2021hierarchical}, and semantic/spatial map representations \cite{Georgakis_2022_CVPR,chen2022weakly}. 
\subsection{Waypoint Prediction}
Waypoint prediction is a widely adopted strategy for VLN-CE. Krantz et al. \cite{Krantz_2021_ICCV} first introduced this framework, and several subsequent works have extended it \cite{krantz2022sim, dcvln, etpnav, smartway}. Typically, waypoint models are pretrained on navigation graphs from the training environments and then kept fixed while the downstream planner is learned. While waypoint prediction has helped close the performance gap in VLN-CE, it still has key limitations. Pretrained models are often trained on limited navigation graphs \cite{dcvln}, restricting generalization; they typically generate only local waypoints and may even predict infeasible ones that collide with obstacles \cite{etpnav}. Moreover, the choice of input representation remains unsettled: An et al. \cite{etpnav} showed that RGB inputs can harm generalization by introducing unnecessary semantics, whereas Shi et al. \cite{smartway} reported improved performance by using stronger RGB-D encoders with an obstacle-awareness loss.

\subsection{Foundation models for VLN}
Foundation models \cite{achiam2023gpt,openai2025gpt5,touvron2023llama,liu2024grounding}, with their strong reasoning and generalization abilities, are particularly well-suited for embodied tasks such as VLN. Existing works have integrated them into navigation pipelines through either zero-shot approaches \cite{navgpt,mapgpt,a2nav,opennav,smartway,aoplanner}, which directly leverage LLMs for decision-making, or fine-tuning strategies \cite{pan2024langnav,Zheng_2024_CVPR,navid}, which adapt open-source vision language models (VLMs) to navigation-specific objectives. 

As for the recent VLN-CE, representative approaches include A2Nav \cite{a2nav}, which decomposes instructions into action-specific subtasks, and waypoint-based methods such as OpenNav \cite{opennav}, which leverages open-source LLMs/VLMs, and SmartWay \cite{smartway}, which integrates a waypoint predictor with an MLLM capable of backtracking. AO-Planner \cite{aoplanner} takes a fully zero-shot route, generating low-level actions from affordance estimation via Grounded SAM \cite{ren2024grounded} and refining candidate paths with an LLM planner. Beyond direct action generation, LLMs have also been explored as value estimators, e.g., InstructNav \cite{long2025instructnav} constructs value maps from textualized observations, while CA-Nav \cite{constraint} enforces sub-instruction constraints to guide navigation.

\section{METHOD}
\subsection{Problem Formulation}
We consider the VLN-CE task within the Habitat simulator, where an embodied agent is equipped with an odometry sensor and an RGB-D camera operating in a 3D mesh environment. At each time step $t$, the agent has access to its current position ($x_t,y_t,z_t$) and a panoramic observation of its surroundings, denoted as $O_t=\{(O_n^{\text{rgb}},O_n^{\text{depth}})\}_{n=1}^{12}$, which consists of twelve RGB and depth image pairs captured at uniformly spaced headings $\{0^\circ,30^\circ,...,330^\circ\}$. The agent’s action space includes low-level controls: \texttt{Move-forward}(0.25m), \texttt{Turn-left/right}($15^\circ$), and \texttt{Stop}. At the beginning of an episode, the agent is provided with a natural language instruction $I=\{w_1,w_2,...,w_N\}$, which implicitly specifies a target location ($x_{\text{target}},y_{\text{target}},z_{\text{target}}$). The goal of the agent is to interpret the instruction and execute a sequence of low-level actions that navigate it successfully to the destination.

\subsection{Waypoint Prediction Model}
\subsubsection{The Abstraction of Input Space} 
Our idea is that both RGB and depth images encode overly complex, high-dimensional information that is not well aligned with the waypoint space, which is fundamentally a 2D radial surface. In particular, RGB inputs often carry semantic-level details that are irrelevant to spatial traversability, while depth images also introduce unnecessary complexity.

In the original waypoint prediction formulation, the space around the agent is discretized into angular intervals of $3^\circ$ and radial steps of $0.25$m, extending up to $3.0$m in each direction. This produces a grid of candidate waypoints represented as a heatmap of size 
$120 \times12$, where each entry corresponds to the likelihood of a waypoint at that location.

To better align the input and output spaces, we propose to construct an obstacle map with the same resolution as the waypoint heatmap. Each cell in this binary map is marked as $1$ if occupied by an obstacle and $0$ if free space. By using this obstacle map as the sole input to the waypoint predictor, we encourage the model to focus exclusively on spatial traversability and to learn the geometric patterns of obstacles and free space surrounding the agent in 2D, ultimately producing more feasible waypoint predictions.

\subsubsection{Obstacle Map Construction} The obstacle map is not directly available and must be constructed from the agent’s depth observations. As is shown in Fig.~\ref{pipeline}, we first convert panoramic depth images into a 3D point cloud and project them onto the discretized $120 \times12$ radial 2D grid used for waypoint prediction. For each grid cell, we estimate local elevation by computing the maximum height of its points, then calculate elevation gradients along each radial direction, treating it as if the agent were “climbing” along that path. A cell is marked as an obstacle if its gradient exceeds a threshold. Unlike prior work \cite{krantz2022sim} that applies a fixed height threshold from the agent’s footpoint, which often misclassifies stairs or slopes as obstacles, our gradient-based method captures such variations more accurately, yielding obstacle maps that better reflect true traversability.

\subsubsection{The Simplified Model Structure} Since our input and output spaces are aligned, we are able to eliminate the complex model architectures used in previous waypoint predictors. Instead, we adopt a lightweight self-attention model with two layers of transformer encoders and a classification head that outputs waypoint logits over the discretized grid. The model is trained using ground-truth waypoint heatmaps as supervision. This lightweighted design not only reduces architectural complexity but also improves waypoint feasibility, ensuring that predicted waypoints are more likely to be located in open and traversable space.

\subsubsection{Masking for linear reachability}To ensure that generated waypoints are truly feasible, they must be compatible with the agent’s low-level motion pattern, which follows a “turn-then-move” strategy. This requires that waypoints be linearly reachable: even if a waypoint lies in open space, it is undesirable if it is located behind an intervening obstacle. To address this, we apply a logit-masking procedure, where for each direction all locations beyond the first detected obstacle are masked out in the waypoint heatmap. From the resulting masked heatmap, we then apply non-maximum suppression (NMS) to select the top $K$ candidate waypoints.

\begin{figure}[t]
    \centering
    \includegraphics[width=0.95\linewidth]{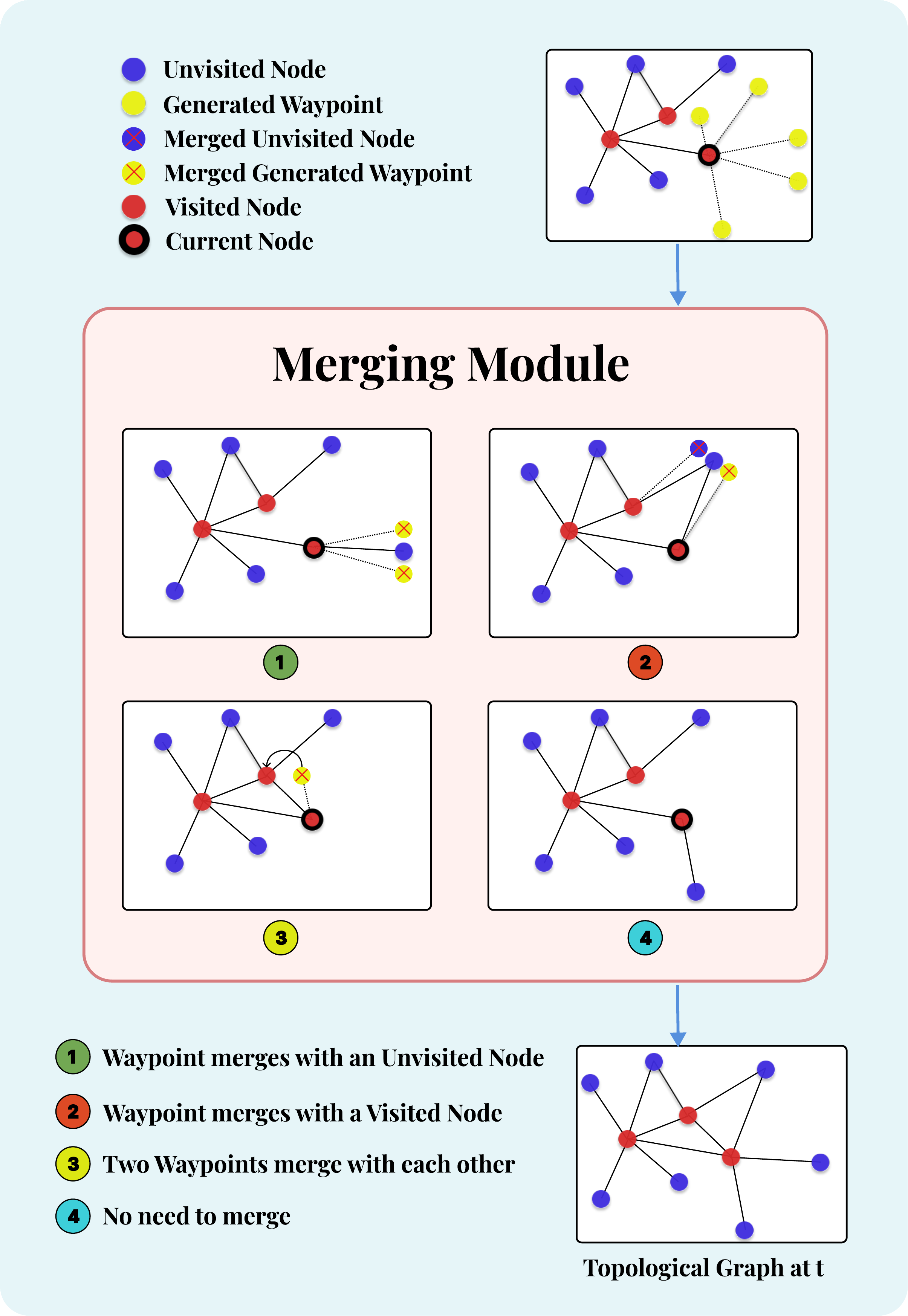}
    \caption{To maintain the topological graph in a simple, effective, and consistent form, the merging module merges a generated waypoint with an existing node, or merges two generated waypoints, if their Euclidean distance falls below a specified threshold. No merging occurs when the generated waypoint lies in an unexplored region.}
    \label{topograph_update}
\end{figure}

\subsection{Topological Graph}
In prior work, the memory mechanisms of LLM-based navigators in continuous environments\cite{smartway, opennav, aoplanner} have been limited to step-wise trajectory history, represented either through visual observations or textual prompts. Such histories record only visit order, without capturing the topological structure of explored areas. As a result, prompts provide merely sequential trajectories, limiting the LLM’s ability to reason about spatial relationships and connectivity. This often leads to two issues: (1) the agent getting stuck or looping in the same area, and (2) the inability to backtrack and re-explore unvisited regions after a navigation error.

Our approach, drawing inspiration from ETPNav \cite{etpnav}, dynamically updates the topological graph by incorporating newly generated waypoints at each time step. The topological graph at each step is denoted as $G_t = (N_t, E_t)$, where $N_t$ denotes the set of all nodes and there exists an edge $e_{ij}\in E_t$ between node $n_i$ and $n_j$ if they are directly reachable via a straight line. The nodes $n_i\in N_t$ are categorized as visited if the agent has already explored them, and as unvisited if they have only been observed but not yet explored. 


    
    
The topological graph update mechanism should maintain \textbf{simplicity}, \textbf{effectiveness}, and \textbf{consistency}, ensuring the MLLM receives a clear spatial representation of explored versus unexplored regions:

\begin{enumerate}
\item \textbf{Avoid regenerating waypoints for visited nodes:} 
When returning to a previously visited node, no new waypoints are generated. Instead, the original waypoints are reused. This ensures that the same action options remain available and preserves a consistent spatial structure within that region.

\item \textbf{Reuse nearby nodes:} Assign new waypoints close in Euclidean distance to existing nodes the same ID and location, preventing clutter and reducing reasoning overhead.
\item \textbf{Remove duplicate waypoints:} Keep only one waypoint among those close in distance, avoiding redundancy and confusion during selection.
\end{enumerate}
\begin{algorithm}[htbp]
\caption{\textsc{GraphUpdate}}
\label{alg:graphupdate}
\DontPrintSemicolon
\KwIn{previous topological graph $G_{t-1}$; current node $n_t$; current observations $o_t$}
\KwOut{Updated graph $G_t$}

$G_t \leftarrow G_{t-1}$\;

\If{$n_t$ is visited}{
    \Return $G_t$
}


$m_t \leftarrow \texttt{get\_obstacle\_map}(o_t)$

$\mathcal{W}_t \leftarrow \texttt{waypoint\_predictor}(m_t)$

$G_t \leftarrow \textsc{MergingModule}(G_t, \mathcal{W}_t, n_t)$;

\Return $G_t$\;
\end{algorithm}


    
Based on these principles, the overall pipeline of the graph update algorithm first determines whether the current location has been visited to decide whether new waypoints should be generated. If new waypoints are required, they are incorporated into the graph through a merging module. The detailed procedures are presented in Algorithm~\ref{alg:graphupdate} and 
Fig.~\ref{topograph_update} illustrates the process of the merging module.

\subsection{MLLM-Based Planner}
 In discrete VLN, methods such as NavGPT \cite{navgpt} and DiscussNav \cite{discussnav} used multi-model pipelines, where VLMs (e.g., BLIP\cite{li2022blip}) generated scene descriptions, LLMs summarized navigation history, and reasoning was performed over textualized inputs. This design suffers from two limitations: compressing visual observations into text loses critical information, and splitting the task across models weakens memory integration. With the advent of MLLMs such as GPT-4o \cite{achiam2023gpt}, recent works like MapGPT \cite{mapgpt} instead leverage raw visual histories and a single MLLM for joint reasoning, while AO-Planner \cite{aoplanner} demonstrated the effectiveness of such prompting in continuous environments. However, history-aware prompting alone cannot fully exploit exploration: it lacks (1) a topological representation of spatial connectivity and (2) explicit tracking of visited/unexplored waypoints. To address this, we propose a TopoGraph-and-VisitInfo–aware prompting system for VLN-CE, which equips the MLLM with structured spatial awareness and enables local path planning and re-exploration of visited regions.

\subsubsection{TopoGraph-and-VisitInfo-Aware Prompting System}
\begin{figure}[tbp!]
    \centering
    \includegraphics[width=0.95\linewidth]{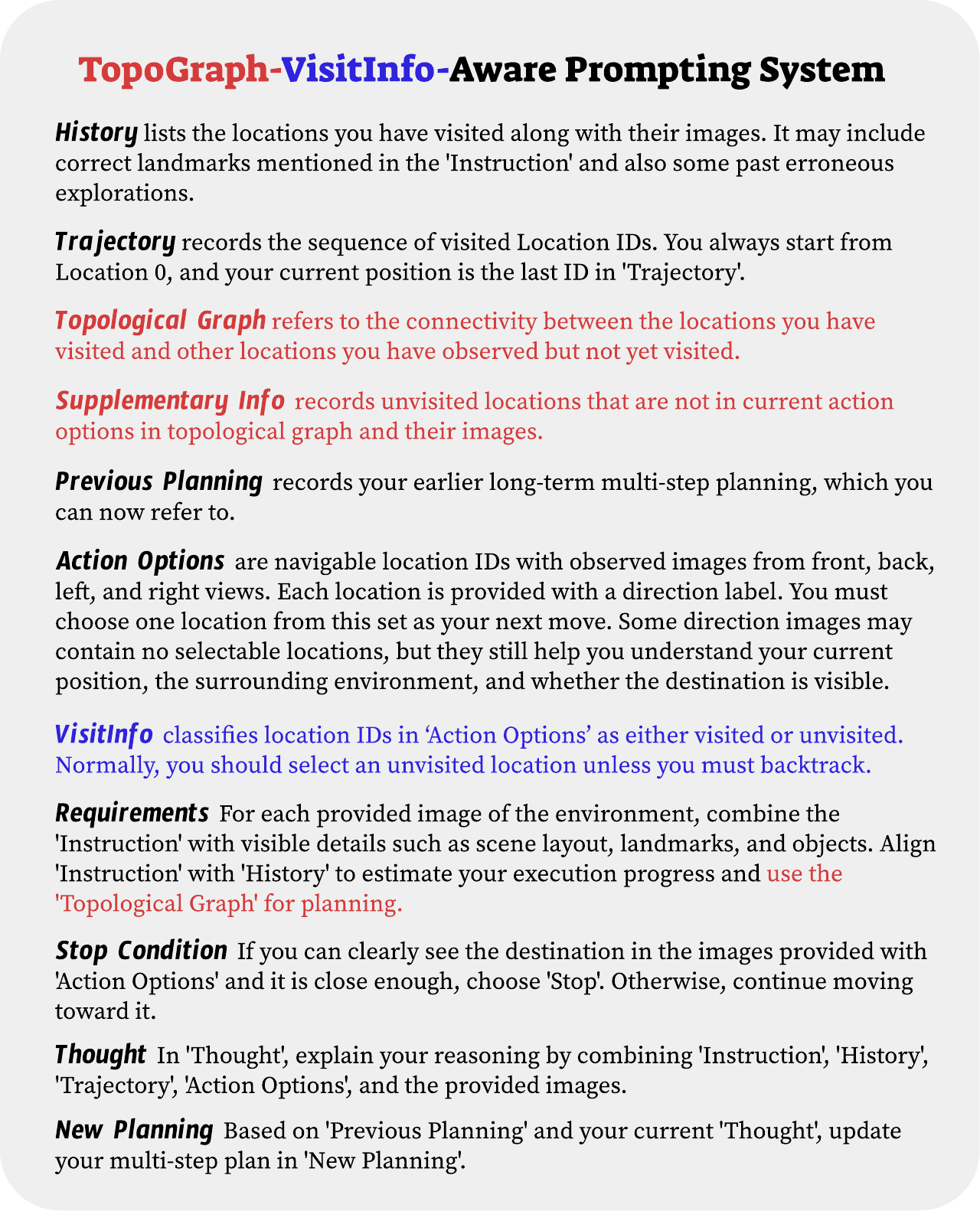}
    \caption{Definition of Each Term in our Prompting System}
    \label{prompting_system}
\end{figure}

Our prompting system receives: (1) Instruction $I$, a global step-by-step guidance; (2) History $H_t$, the sequence of visited locations visualized in their corresponding images; (3) Trajectory $Tr_t$, the sequence of visited location IDs; (4) Topological Graph $G_t$, the connectivity between nodes; (5) VisitInfo $V_t$, the visitation status of the current action options; (6) Supplementary Info $S_t$, locations that are unvisited and their corresponding images. (7) Action Options $A_t$, navigable location IDs with observed images from the front, back, left, and right views. The output of our prompting system are Thought $T_t$ and Action $a_t$. The detailed definition of each term in our prompting system are listed in Fig.~\ref{prompting_system}, and it serves as the system prompt. Unlike AO-Planner\cite{aoplanner}, which projects RGB pixels into 3D using depth images and camera intrinsics, we instead use predicted 3D waypoint positions with camera intrinsics to project them back onto the RGB images. To incorporate the topological graph into the prompting system, we follow the approach of MapGPT\cite{mapgpt} by representing the connectivity of each visited node to others through natural language descriptions. VisitInfo can be directly obtained from the topological graph by the distinction between visited and unvisited nodes.

\begin{equation*}
    T_t, a_t = MLLM(I, H_t, Tr_t, G_t, V_t, S_t, A_t)
\end{equation*}

\subsubsection{Encourage Exploration and Equip MLLMs with Local Path Planning}
The Topograph-and-VisitInfo-aware prompting system alleviates two common challenges faced by zero-shot methods in the VLN-CE setting. First, prior approaches do not record visitation information, which often causes the agent to repeatedly explore the same region without awareness, ultimately becoming trapped in a confined loop. By incorporating visit information, our system encourages exploration by prioritizing unvisited locations as candidate targets, unless a navigation error occurs and the MLLM determines that backtracking to previously visited locations is necessary. We observe that this exploration-oriented strategy effectively mitigates the issue of the agent getting stuck in a limited area.

\begin{figure}[t!]
    \centering
    \includegraphics[width=0.95\linewidth]{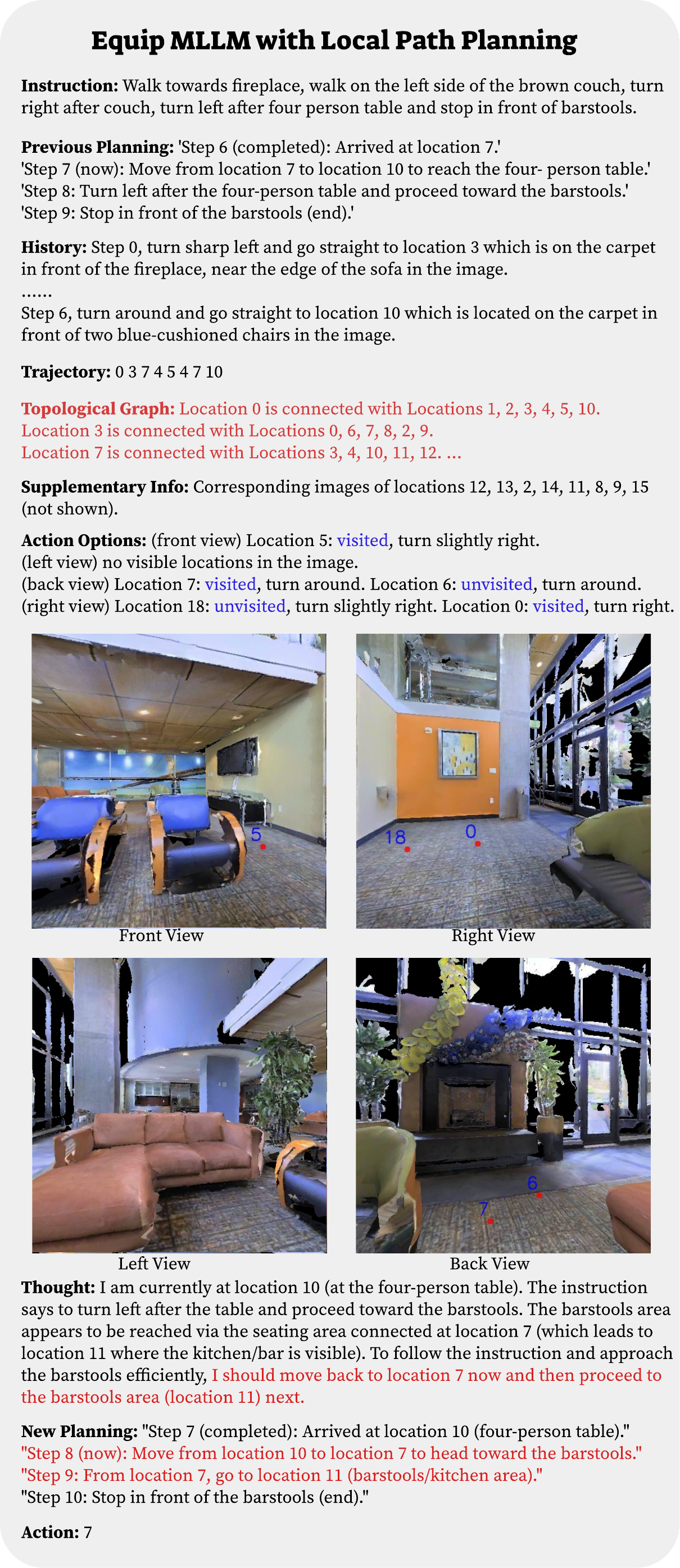}
    \caption{An example of the complete prompting information at a single step. In this scenario, a navigation error occurs, and the MLLM leverages the connectivity information in the topological graph to perform local path planning for correction: it first backtracks to a previously visited location and then selects an unvisited node that is better aligned with the instruction.}
    \label{reasoning-process}
\end{figure}
Second, when navigation errors arise, earlier prompting systems fail to provide sufficient information for the MLLM to conduct path planning for error correction. Our system addresses this by supplying the topological graph as a source of spatial structural knowledge. This allows the MLLM to utilize connectivity information between locations for planning, such as backtracking to a previously visited node and then exploring other interested unvisited locations, or even retreating to earlier visited locations to reconsider exploration strategies. We find that this capability of local path planning enables the MLLM to timely correct its route when navigation errors occur, thereby preventing large deviations from the instruction-guided trajectory. Fig.~\ref{reasoning-process} illustrates the complete prompting information at a single step, as well as how the MLLM leverages the topological graph to perform local path planning for error correction.

\section{Experiments}
\subsection{Experimental Settings}
\noindent \textbf{Dataset} We evaluate our navigator on R2R-CE \cite{r2rce} and RxR-CE \cite{rxr}. R2R-CE extends the R2R dataset \cite{r2r} into continuous environments within the Habitat simulator \cite{habitat}. RxR-CE is more challenging, with longer instructions, longer trajectories, and no sliding along obstacles, increasing the chance of getting stuck. We compare our method against both learning-based and LLM/MLLM-based navigators. For R2R-CE, evaluation is conducted on the full val-unseen split (1,839 episodes across 11 scenes). For RxR-CE, following prior work \cite{a2nav,aoplanner,constraint}, we use 300 sampled episodes from the English val-unseen split. To reduce API costs, we also sample 300 episodes for ablation studies. All experiments use the GPT-5-mini API \cite{openai2025gpt5} as the MLLM.

We train our waypoint predictor on the train split of ground-truth connectivity graph used in DC-VLN\cite{dcvln}, which is adapted from the pre-defined graph for MP3D\cite{chang2017matterport3d} to fit the continuous environments in Habitat. We use the AdamW\cite{adamw} optimizer with a learning rate of $10^{-4}$, a batch size of 32, and 30 training epochs. Evaluation is performed on the val-unseen split of the ground-truth connectivity graph.

\noindent \textbf{Evaluation Metrics} For the evaluation of our navigator in R2R-CE task, we use the standard metrics used in VLN-CE: navigation error (NE), oracle success rate (OSR), success rate (SR), success weighted by path length (SPL) and normalized dynamic-time warping (nDTW). 

For the evaluation of our waypoint predictor, we adopt the same metrics as DC-VLN \cite{dcvln}. Specifically, $|\Delta|$ denotes the difference between the number of predicted waypoints and the ground-truth targets, \%Open represents the proportion of predicted waypoints located in open space, and Avg\_Score represents the score of the generated waypoints on the ground-truth heatmap, where waypoints closer to the ground-truth waypoint receive higher scores. In addition, $d_C$ and $d_H$ correspond to the Chamfer distance and Hausdorff distance, respectively, both of which are standard measures for quantifying discrepancies between point clouds.

\subsection{Experimental Results}

\begin{table}[t]
\centering
\small
\caption{Comparison of different methods on the val-unseen split of R2R-CE in Habitat simulator}
\label{tab:r2rce}
\begin{tabular}{@{}lcccc@{}}
\toprule
\textbf{Method} &
\textbf{NE}\(\downarrow\) &
\textbf{OSR}\(\uparrow\) &
\textbf{SR}\(\uparrow\) &
\textbf{SPL}\(\uparrow\) \\
\midrule
\multicolumn{5}{c}{\bfseries Supervised} \\
\midrule
Seq2Seq\cite{r2rce}       & 7.77 & 37   & 25   & 22   \\
CMA\cite{r2rce}           & 7.37 & 40   & 32   & 30   \\
DC-VLN\cite{dcvln}        & 5.89 & 51   & 42   & 36   \\
Navid\cite{navid}         & 5.47 & 49   & 37   & 36   \\
ETPNav\cite{etpnav}       & 4.71 & 65   & 57   & 49   \\
\midrule
\multicolumn{5}{c}{\bfseries Zero-Shot} \\
\midrule
OpenNav\cite{opennav}     & 6.70 & 23   & 19   & 16.1 \\
A2Nav\cite{a2nav}         & --   & --   & 23   & 11.1 \\
AO-Planner\cite{aoplanner}& 6.95 & 38   & 25   & 16.6 \\
CA-Nav\cite{constraint}   & 7.58 & 48   & 25   & 10.8 \\
SmartWay\cite{smartway}   & 7.01 & 51   & 29   & 22.0 \\
\textbf{Ours}             & \textbf{6.12} & \textbf{55} & \textbf{41} & \textbf{25.4} \\
\bottomrule
\end{tabular}
\end{table}

\begin{table}[t]
\centering
\small
\caption{Comparison of different methods on the val-unseen split of RxR-CE in Habitat simulator}
\label{tab:rxrce}
\begin{tabular}{@{}lcccc@{}}
\toprule
\textbf{Method} &
\textbf{NE}\(\downarrow\) &
\textbf{SR}\(\uparrow\) &
\textbf{SPL}\(\uparrow\) &
\textbf{nDTW}\(\uparrow\) \\
\midrule
\multicolumn{5}{c}{\bfseries Supervised} \\
\midrule
Seq2Seq\cite{r2rce}   & 12.10 & 13.9 & 11.9 & 30.8 \\
DC-VLN\cite{dcvln}    & 8.98  & 27.1 & 22.7 & 46.7 \\
Navid\cite{navid}     & 8.41  & 23.8 & 21.2 & --   \\
ETPNav\cite{etpnav}   & 5.64  & 54.8 & 44.9 & 61.9 \\
\midrule
\multicolumn{5}{c}{\bfseries Zero-Shot} \\
\midrule
A2Nav\cite{a2nav}          & --    & 16.8 & 6.3  & --   \\
AO-Planner\cite{aoplanner} & 10.75 & 22.4 & 15.1 & 33.1 \\
CA-Nav\cite{constraint}    & 10.37 & 19.0 & 6.0  & 13.5 \\
\textbf{Ours}              & \textbf{7.56} & \textbf{35.7} & \textbf{21.7} & \textbf{42.4} \\
\bottomrule
\end{tabular}
\end{table}

\setlength{\tabcolsep}{4pt}
\begin{table}[t]
\centering
\small
\caption{Comparison of different waypoint predictors on val-unseen split of MP3D ground truth graph}
\label{tab:waypoint}
\begin{tabular}{@{}p{2.5cm}ccccc@{}}
\toprule
 \textbf{Model} & \(|\Delta|\) & \%Open \(\uparrow\) &  Avg\_Score \(\uparrow\) & \(d_c \downarrow\) & \(d_h \downarrow\) \\
\midrule
DC-VLN$^{\dagger\star}$\cite{dcvln} & 1.40 & 79.86 & 1.30 & 1.07 & 2.00 \\
ETPNav$^{\star}$\cite{etpnav} & 1.39 & 84.05 & 1.40 & 1.04 & 2.01 \\
SmartWay$^{\dagger\star}$\cite{smartway} & 1.41 & 87.26 & 1.48 & 1.03 & \textbf{1.96} \\

\textbf{Ours}$^{\ddagger}$ & 1.41 & \textbf{90.18} & \textbf{1.56} & \textbf{1.02} & 2.00 \\
\bottomrule
\end{tabular}
\vspace{1ex}
\raggedright \footnotesize
$^{\dagger}$ RGB as input. \quad 
$^{\star}$ Depth as input. \quad 
$^{\ddagger}$ Obstacle map as input.
\end{table}

Table~\ref{tab:r2rce} reports the results on R2R-CE. As expected, supervised methods generally outperform zero-shot ones due to task-specific training. Our method, however, establishes a new state-of-the-art among zero-shot approaches, showing consistent improvements across all metrics. Notably, its performance is already comparable to the average of supervised methods, narrowing the gap to strong baselines such as DC-VLN and reducing the difference with the SOTA supervised method ETPNav.

Table~\ref{tab:rxrce} shows the results on RxR-CE. Our method establishes a new state-of-the-art in the zero-shot setting, with substantial gains over previous zero-shot SOTAs. Its performance is also comparable to several supervised methods, though a noticeable gap remains compared to ETP-Nav, the strongest supervised model, particularly on SR and nDTW.

Our abstract obstacle-based waypoint predictor is significantly more lightweight and easier to train than prior models, which often rely on complex cross-attention over RGB-D features and require hundreds of training epochs. In contrast, ours can be trained in only 30 epochs. As shown in Table~\ref{tab:waypoint}, it achieves the best results on \%Open, Avg\_Score, and $d_C$. Notably, \%Open is improved by 3\%, which indicates that the obstacle map effectively encourages the model to generate waypoints within free space. These results further suggest that semantic information from RGB inputs is unnecessary—while redundant details in depth images can be compactly captured by the obstacle map.

\subsection{Ablation Studies}
\begin{table}[t]
\centering
\small
\caption{Ablation experiments of our prompting system on the val-unseen split of R2R-CE}
\label{tab:prompting}
\begin{tabular}{@{}lcc@{}}
\toprule
\textbf{Prompting System} & \textbf{SR$\uparrow$} & \textbf{SPL$\uparrow$} \\
\midrule
Original         & 46 & 27.5 \\
w/o VisitInfo    & 40 & 22.7 \\
w/o TopoGraph    & 37 & 22.6 \\
\bottomrule
\end{tabular}
\end{table}
\noindent \textbf{MLLM-based-Planner} To evaluate the effectiveness of the topological graph and visit information in our prompting system, we conduct ablation studies using three variants: the proposed TopoGraph-and-VisitInfo-Aware prompting system, a version without the topological graph, and a version without visit information. Table~\ref{tab:prompting} shows the results of ablation experiments of our prompting system. The absence of visit information leads to a 6\% drop in SR, and a 4.8\% drop in SPL. The absence of topological graph results in an 9\% decrease in SR and a 4.9\% decrease in SPL. These results highlight the effectiveness of the topological graph for capturing spatial structure and the value of visit information for identifying explored areas.

\begin{table}[t]
\centering
\small
\caption{Ablation experiments of waypoint predictors using our prompting system on the val-unseen split of R2R-CE}
\label{tab:ablationwp}
\begin{tabular}{@{}lcccc@{}}
\toprule
\textbf{Waypoint Predictor} & \textbf{SR$\uparrow$} & \textbf{SPL$\uparrow$} & \textbf{Collision$\downarrow$} \\
\midrule
Ours     &  46 & 27.5 & 3.85\\
DC-VLN   & 42 & 25.4 & 9.98\\
\bottomrule
\end{tabular}
\end{table}
\noindent \textbf{Waypoint Predictor} 
We compare the performance of our waypoint predictor with  DC-VLN under the R2R-CE task. For fairness, both methods are evaluated using our prompting system. Table~\ref{tab:ablationwp} shows the results of ablation experiments of different waypoint predictors. If the waypoint predictor is changed to DC-VLN, SR decreases by 4\%, SPL decreases by 2.1\% and Collision Rate increases by 6.13\%. This indicates that our waypoint predictor is more likely to generate waypoints in open space and achieves better alignment with the ground-truth connectivity graph.

\noindent \textbf{Comparison Under the Same API} 
\begin{table}[t]
\centering
\small
\caption{Comparative experiments under the same API on the val-unseen split of R2R-CE}
\label{tab:sameapi}
\begin{tabular}{@{}lccc@{}}
\toprule
\textbf{Waypoint Predictor} &\textbf{Prompting System} & \textbf{SR$\uparrow$} & \textbf{SPL$\uparrow$} \\
\midrule
Ours        & Ours-GPT5mini  & 46 & 27.5 \\
AO-planner   & AO-planner-GPT5mini & 33 & 12.7 \\
Ours   & AO-planner-GPT5mini & 38 & 19.9 \\
\bottomrule
\end{tabular}
\end{table}
For fairness, we adapt the SOTA model AO-Planner by replacing its high-level agent API with GPT-5-mini and substituting its low-level planner with our waypoint predictor. The results of comparative experiments under the same API are shown in Table~\ref{tab:sameapi}. Comparing the first and second rows, we observe that with the same API, our pipeline still achieves a substantial improvement, with SR higher by 13\% and SPL higher by 14.8\% than AO-Planner. This gain comes from both our obstacle-based waypoint predictor and our TopoGraph-and-VisitInfo-aware prompting system. Comparing the first and third rows, we find that when using the same waypoint predictor, our prompting system improves SR by 8\% and SPL by 7.6\%. Finally, comparing the second and third rows, we observe that simply replacing AO-Planner’s waypoint predictor with ours yields an improvement of 5\% in SR and 7.2\% in SPL.
\section{CONCLUSIONS}

We proposed a zero-shot VLN-CE framework that combines a simplified waypoint predictor with an MLLM-based planner. The waypoint predictor operates on an abstract obstacle map, ensuring linearly reachable waypoints. We further introduced a topological graph with visitation records, incorporated into the MLLM prompt alongside visual inputs: the graph encodes spatial structure, while visitation records capture exploration history. This design enables the MLLM to reason more effectively about space and progress, encouraging exploration and enabling local path planning to backtrack and re-explore for error correction. Experiments on R2R-CE and RxR-CE show that our framework outperforms prior state-of-the-art zero-shot methods and achieves performance competitive with supervised approaches.

\addtolength{\textheight}{-1cm}   









\bibliographystyle{IEEEtran}
\bibliography{ref}

\end{document}